%% file: main.tex
\documentclass{article}

\usepackage{microtype}
\usepackage{graphicx}
\usepackage{subcaption}
\usepackage{booktabs} 

\usepackage{hyperref}

\usepackage[preprint]{icml2026}

\usepackage{amsmath}
\usepackage{amssymb}
\usepackage{mathtools}
\usepackage{amsthm}

\usepackage[capitalize,noabbrev]{cleveref}

\theoremstyle{plain}

\theoremstyle{definition}

\theoremstyle{remark}

\usepackage[textsize=tiny]{todonotes}

\icmltitlerunning{Fault-Tolerant Evaluation for Sample-Efficient Model Performance Estimators}

\input{def}

\begin{document}

\twocolumn[
  \icmltitle{Fault-Tolerant Evaluation for Sample-Efficient Model Performance Estimators}



  \icmlsetsymbol{equal}{*}

  \begin{icmlauthorlist}
    \icmlauthor{Zihan Zhu}{mqu}
    \icmlauthor{Yanqiu Wu}{mqu}
    \icmlauthor{Qiongkai Xu}{mqu}
  \end{icmlauthorlist}

  \icmlaffiliation{mqu}{School of Computing, FSE, Macquarie University, Sydney, Australia}

  \icmlcorrespondingauthor{Qiongkai Xu}{qiongkai.xu@mq.edu.au}

  \icmlkeywords{Machine Learning, ICML}

  \vskip 0.3in
]



\printAffiliationsAndNotice{}  

\begin{abstract}

  In the era of Model-as-a-Service, organizations increasingly rely on third-party AI models for rapid deployment. However, the dynamic nature of emerging AI applications, the continual introduction of new datasets, and the growing number of models claiming superior performance make efficient and reliable validation of model services increasingly challenging. This motivates the development of \textit{sample-efficient performance estimators}, which aim to estimate model performance by strategically selecting instances for labeling, thereby reducing annotation cost. Yet existing evaluation approaches often fail in low-variance settings: RMSE conflates bias and variance, masking persistent bias when variance is small, while $p$-value based tests become hypersensitive, rejecting adequate estimators for negligible deviations. To address this, we propose a \textit{fault-tolerant evaluation framework} that integrates bias and variance considerations within an adjustable tolerance level $\varepsilon$, enabling the evaluation of performance estimators within practically acceptable error margins. We theoretically show that proper calibration of $\varepsilon$ ensures reliable evaluation across different variance regimes, and we further propose an algorithm that automatically optimizes and selects $\varepsilon$. Experiments on real-world datasets demonstrate that our framework provides comprehensive and actionable insights into estimator behavior.


\end{abstract}

\begin{figure*}[ht]
  \begin{center}
    \centerline{\includegraphics[width=0.9\linewidth]{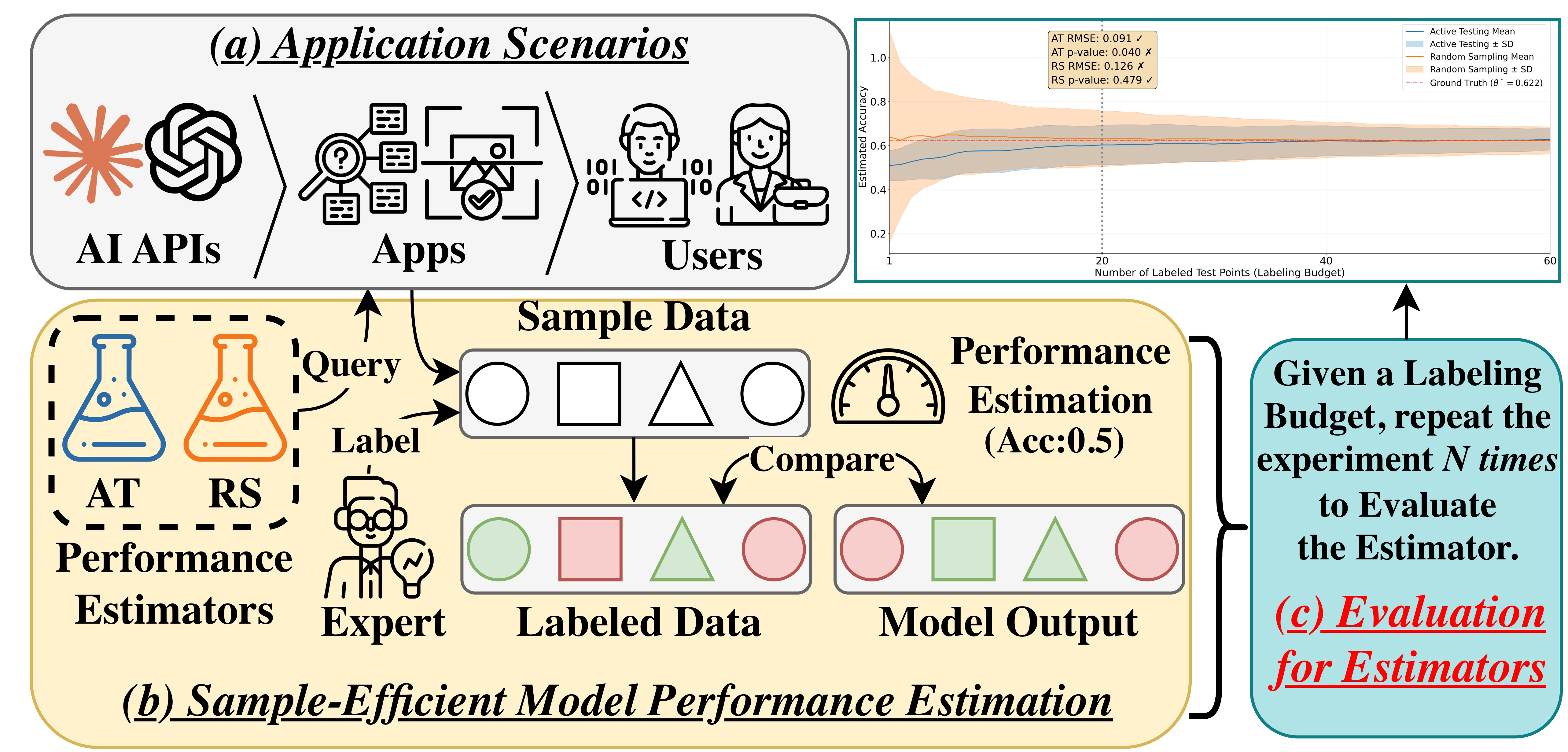}}
    \caption{
      An overview of the evaluation challenge for sample-efficient model performance estimators. (a) AI models accessed via web APIs support various applications and users. (b) Performance estimators (\eg Active Testing or Random Sampling) query and label task samples within a labeling budget to estimate model performance. (c) \textit{Estimator evaluation} (our contribution): The full estimation process is repeated $N$ times to assess estimator quality against ground truth.}
    \label{fig:teaser}
  \end{center}
  \vspace*{-2em}
\end{figure*}

\section{Introduction}


In the modern web ecosystem, AI models are increasingly accessed as services via APIs and embedded into agentic systems that directly support downstream decision making. The scale of adoption is striking: by 2024, global corporate AI investment reached \$252.3 billion, and 78\% of organizations reported active AI usage \cite{maslej2025artificial}. With this rapid growth, organizations routinely compare third-party models--ranging from traditional classifiers like ResNet \cite{he2016deep} and BERT \cite{devlin2019bert} to large language models such as GPT-4o \cite{hurst2024gpt} and Claude Sonnet 4 \cite{anthropic2025system}--to determine whether adopting, upgrading, or replacing a model is economically and operationally justified. Reliable performance estimation is therefore critical, particularly in settings where models are deployed under rapidly changing data distributions and labeling high-quality test data is costly or infeasible \cite{maslej2025artificial}. 

To address the scarcity of labeled data, sample-efficient performance estimation has emerged as a practical solution. These corresponding estimators strategically select a small yet most informative set of instances for annotation in order to estimate model performance with reduced labeling cost. Methods such as \textit{Active Testing} have demonstrated substantial variance reduction and faster convergence compared to naive performance estimation strategies like \textit{Random Sampling}. Despite significant progress in designing sample-efficient estimators, a further fundamental question remains insufficiently addressed: \textit{How should we evaluate the quality of performance estimators?}

Most prior work \cite{nguyen2018active, kossen2021active, ji2021active} typically evaluates the quality of estimators using point-wise metrics such as MSE or RMSE, or via statistical significance tests that assess whether an estimator’s mean differs from ground truth. While widely adopted, these approaches exhibit fundamental limitations in low-variance regimes, often leading to misleading conclusions. MSE and RMSE aggregate bias and variance into a single measure, which can obscure persistent systematic bias once variance is sufficiently reduced. $p$-value–based significance tests provide a complementary perspective by explicitly testing deviation from the ground truth, but become increasingly sensitive as variance decreases, often rejecting estimators that are practically accurate.



These limitations manifest directly in practical evaluation workflows. Figure~\ref{fig:teaser} illustrates how they arise in the evaluation of sample-efficient model performance estimators and situates our contribution within this workflow. In practice, AI models accessed through web APIs are used across a variety of applications, each requiring reliable information about model performance. To obtain this information under a limited labeling budget, performance estimators such as Active Testing (AT) and Random Sampling (RS) query task instances, obtain expert labels, and compare model outputs to produce an estimated accuracy. Panel (c) highlights the core challenge addressed in this paper: traditional metrics such as RMSE and $p$-values often provide conflicting signals about estimator quality. This contradiction arises from inherent limitations in how traditional metrics handle the bias-variance trade-off (Section~\ref{sec:limit_trad}), and our results show that such failures occur consistently across datasets and labeling budgets (Section~\ref{sec:conflict_signal}). Consequently, practitioners are uncertain about which estimator to trust, leading to suboptimal model choices and inefficient use of labeling resources.

To address these limitations, we introduce a fault-tolerant evaluation (FT-Eval) framework that allows users to specify an acceptable tolerance level $\varepsilon$ when evaluating the performance of estimators. Our framework employs two bounded one-sided $t$-tests (TOST), inspired by \cite{schuirmann1987comparison}, to evaluate whether estimators' outputs fall within a user-defined tolerance interval around the ground truth (detailed in Section~\ref{sec:method}). We further show theoretically that proper calibration of $\varepsilon$ yields reliable evaluation across different variance regimes, and we propose an algorithm that automatically selects $\varepsilon$. Overall, \ourframework is broadly applicable to settings where reliable model evaluation must be achieved under limited labeling budgets, including API-based AI services and large-scale agentic AI systems.


The main contributions of our work are as follows:
\begin{itemize}
    \item We identify the fundamental limitations of traditional evaluation approaches (RMSE and $p$-values) for assessing performance estimators, and reveal their causes.
    
    \item We propose a novel fault-tolerant evaluation framework that assesses estimator quality within user-defined tolerance levels, explicitly incorporating both bias and variance considerations.

    \item We design a practical algorithm that automatically select and optimize tolerance levels in accordance with the available labeling budget.
    
    \item We validate our framework through extensive experiments on various AI models, datasets, and performance estimators--we show that traditional metrics provide conflict guidance in 73\% cases while \ourframework yields consistent, interpretable, and actionable evaluations. 
\end{itemize}

\section{Related Work}

\subsection{Active Learning}
Although unlabeled data are often abundant, acquiring labels can be costly. Active learning \cite{atlas1989training, settles2009active} approaches this challenge by allowing learning algorithms to selectively query the most informative instances to be labeled by an oracle (e.g., a human annotator), thereby achieving higher accuracy with fewer training labels. Various query strategies have been developed under this paradigm. For example, uncertainty sampling \cite{lewis1994heterogeneous,lewis1995sequential} selects instances for which the model is least confident, and has been widely applied in text classification and sequence labeling tasks. The query-by-committee approach \cite{seung1992query} maintains a diverse set of hypotheses and selects queries with maximal disagreement among them, offering theoretical guarantees under the version space framework. Additional strategies include expected error reduction \cite{roy2001toward}, which selects instances expected to minimize future generalization error, and expected model change \cite{settles2007multiple}, which selects queries likely to induce the largest parameter updates.

Recent research has particularly focused on deep active learning (DAL), which adapts AL principles to deep neural networks. DAL methods address unique challenges such as dynamically changing feature representations during training and the need to leverage pre-trained models effectively \cite{li2024survey}. Recent advances have particularly emphasized the integration of DAL with large-scale pre-trained language models, where selecting just a few of labeled samples for fine-tuning can achieve competitive performance compared to full dataset training \cite{maekawa2022low, seo2022active}.

\subsection{Sample-Efficient Performance Estimation}
Sample-efficient performance estimation aims to accurately approximate a model’s performance while minimizing labeling costs. A prominent paradigm within this domain is Active Testing (AT). Unlike AL, which focuses on selecting data to improve model training, AT aims to construct an efficient and accurate performance estimate of the model by strategically selecting informative instances to label.

\citet{nguyen2018active} propose an AT framework for efficiently and robustly estimating model performance on large-scale visual recognition datasets with noisy labels. However, the framework assumes access to initial noisy annotations and focus on correcting them through selective vetting, which introduces substantial bias. \citet{kossen2021active} extend this paradigm by introducing acquisition functions based on uncertainty estimates from surrogate models. These models are trained jointly on observed test data and original training data, and are iteratively refined as more test labels are acquired. Their approach provides unbiased test loss estimates with significantly reduced variance, but it depends on retraining surrogate models throughout the evaluation process, which can be computationally intensive.

\citet{ji2021active} introduce a lightweight active Bayesian assessment framework for evaluating black-box classifiers. Their method leverages Thompson sampling with Beta-Bernoulli conjugation to estimate group-wise accuracy. A key aspect of their framework is its reliance on the predicted class labels and confidence scores generated by the classifier for each test data point. These predictions are used to partition the dataset into distinct groups and initialize informative priors, a process that is computationally feasible given the relatively low inference costs of traditional classifiers.

Recent work such as AcTracer \cite{huang2024active} introduces an AT framework for large language models (LLMs). AcTracer adopts a multi-stage sampling strategy that utilizes both internal features (e.g., intermediate representations) and external outputs (e.g., confidence scores) to guide label acquisition. Furthermore, \citet{tahan2024label} propose DiffUse, a label-efficient method for model selection in text generation tasks. By clustering semantic difference vectors between model outputs, DiffUse identifies diverse and representative examples for oracle labeling.

Despite the diversity of AT strategies, most existing work in this space evaluates them using standard metrics such as Root Mean Squared Error (RMSE) or MSE. However, these traditional metrics provide an incomplete assessment of AT estimator quality. They conflate bias and variance into a single value, making it difficult to distinguish whether errors arise from systematic bias or random variability. To address these limitations, we propose a fault-tolerant evaluation framework that allows users to specify acceptable tolerance levels and provides a more comprehensive assessment of estimator performance.

\section{Problem Formulation and Motivation}
We formulate the problem of evaluating model performance estimators and review traditional evaluation methods. Then, we discuss their inherent limitations, particularly the issue associated with variance reduction that naturally arises as the labeling budget increases or as more advanced estimation strategies are applied.

\subsection{Preliminary}

\paragraph{Model Performance Estimation} Given a machine learning model or large language model $\mathcal{M}$ and a dataset $\mathcal{D}$ associated with a specific application, the primary goal is to estimate the performance of $\mathcal{M}$ on $\mathcal{D}$. We denote this true performance by $\theta^*$ (\eg overall accuracy on the whole dataset). In practice, $\theta^*$ cannot be directly observed before exhaustively querying and testing all instances in $\mathcal{D}$. 
Sample-efficient performance estimators aim to estimate $\theta^*$ using only a selective subset $\mathcal S \subseteq \mathcal D$. The annotation procedure proceeds sequentially, gradually forming the labeled set $\mathcal S$ under a labeling budget $|\mathcal{S}|$.\footnote{Throughout this paper, we use the terms \textit{labeling budget} and \textit{number of labeled test samples} interchangeably to describe $|\mathcal{S}|$.}

\paragraph{Evaluating Estimation Methods} Each estimator is evaluated by repeating the annotation procedure $N$ times with different random seeds, producing a collection of $N$ independent performance estimates $(\hat{\theta}_1, \hat{\theta}_2, \ldots, \hat{\theta}_N)$, which form the basis of our statistical evaluations in later sections. There are two standard metrics for evaluating the estimators: the root mean squared error (RMSE) and the $p$-value from a traditional two-sided $t$-test. These metrics serve as baselines for deciding which estimator to adopt in practice.

\begin{figure*}[h!]
    \centering
    \includegraphics[width=\linewidth]{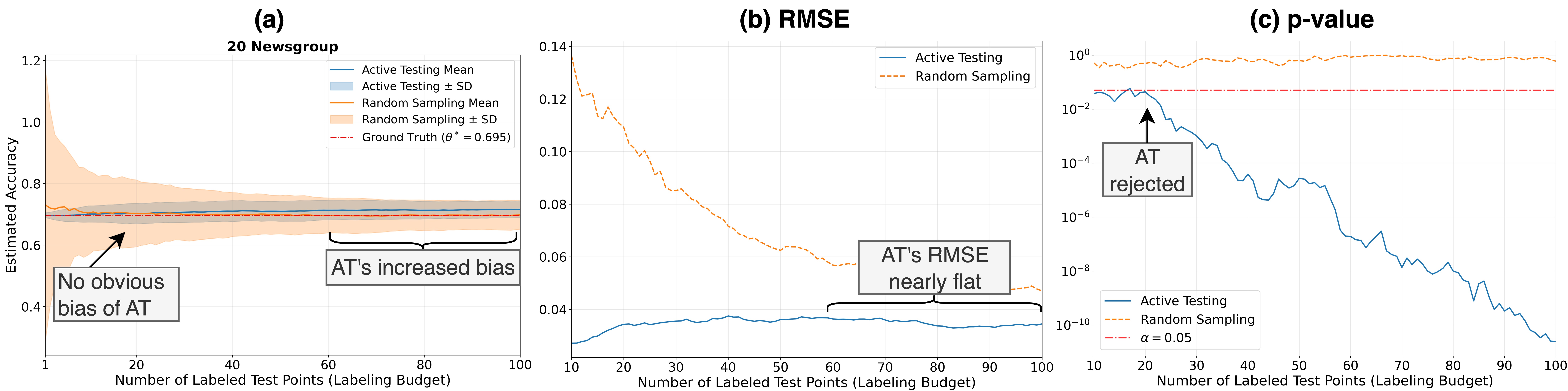}
    \caption{A comparison on two estimators, active testing (AT) and random sampling (RS), on 20 Newsgroup: (a) estimated performance (\ie accuracy) with their mean and standard deviation across multiple runs, against the ground truth performance $\theta^* = 0.695$ (the red dashed line); (b) RMSE, and (c) $p$-values from the traditional two-sided $t$-test.} 
    \label{fig:limit_p_rmse_example}
    \vspace{-1em}
\end{figure*}

\subsection{Limitation of Traditional Metrics}
\label{sec:limit_trad}

We identify key pitfalls of prevalent evaluation methods under \textit{low-variance} conditions. First, metrics such as MSE and RMSE combine bias and variance into a single scalar, causing their values to remain small even when systematic bias persists as variance decreases. Second, statistical significance testing (\eg via $p$-values) becomes unstable in low-variance regimes, where even minor biases are amplified, leading to potentially misleading conclusions about estimator quality.



Moreover, low variance naturally arises in two common cases: (1) when the labeling budget increases, the decrease of variance is grounded by the large number theorem~\cite{kolmogorov1963theory}; and (2) advanced performance estimators (\eg active testing) tend to achieve lower variance at earlier stages. 
The following section provides the theoretical analysis of each metric and demonstration with practical evidence.

\subsubsection{Mean Squared Error (MSE) and Root Mean Squared Error (RMSE)}
Traditional point estimation metrics, such as MSE and RMSE \cite{fisher1922mathematical}, are computed as follows:
\begin{equation}
\vspace{-0.3em}
\text{MSE} = \frac{1}{N} \sum_{i=1}^{N} (\hat{\theta}_i - \theta^*)^2, \quad \text{RMSE} = \sqrt{\text{MSE}}\,,
\end{equation}
where $\hat{\theta}_i$ denotes the $i$-th estimate obtained over $N$ independent runs, and $\theta^*$ is the ground truth. The MSE can be decomposed into bias and variance components,
\begin{equation}
\text{MSE} = (\bar{\theta} - \theta^*)^2 + \frac{1}{N} \sum_{i=1}^{N} (\hat{\theta}_i - \bar{\theta})^2 = \text{Bias}^2 + \text{Var},
\label{eq:mse_decomposition}
\end{equation}
where $\bar{\theta}$ is the average of $\{\hat{\theta_i}\}$.
\paragraph{Limitations of MSE and RMSE}
While MSE and RMSE are widely used metrics, they exhibit critical limitations when evaluating performance estimators across varying labeling budgets. Specifically, the inevitable reduction in variance creates a fundamental problem: the variance component can drive (R)MSE arbitrarily low even when systematic bias remains constant or worsens. Consequently, a biased estimator can achieve misleadingly low (R)MSE values by suppressing variance through larger budgets or variance-reduction strategies such as active testing, even when systematic accuracy does not improve.


This behavior is problematic for two reasons. First, (R)MSE conflates two fundamentally different sources of error: systematic deviation from the truth (bias) and random fluctuation across runs (variance). An estimator with zero bias but moderate variance may be preferable to one with low variance but substantial bias, yet (R)MSE treats these scenarios identically if they yield the same squared error. Second, since variance naturally decreases with budget size while bias may remain unchanged or even increase, (R)MSE can give a misleading impression of estimator quality by rewarding budget increases that merely suppress variance without addressing systematic errors.

Figure~\ref{fig:limit_p_rmse_example} illustrates this limitation using Active Testing (AT) and Random Sampling (RS) for estimating performance of BERT classifiers on 20 Newsgroup. Plot~(a) shows the estimated accuracy means and sample standard deviations across runs, while plot~(b) presents the corresponding RMSE values. Notably, AT's RMSE remains relatively stable across the 10-100 labeling budget range, fluctuating only slightly between 0.034 and 0.036. This stability suggests consistent performance and might lead practitioners to conclude that AT maintains its accuracy throughout. However, plot (a) reveals a systematic upward bias emerging in the 60-100 point range, where AT's mean estimates consistently overestimate the ground truth $\theta^* = 0.695$. Despite this growing systematic error, RMSE fails to reflect the performance of AT in this case because the continuing variance reduction compensates for the bias increase in the overall RMSE calculation.

\subsubsection{Traditional Two-Sided Test}
The traditional statistical significance test \cite{neyman1933ix} evaluates the quality of an estimator by assessing whether its sample mean $\bar{\theta}$ differs significantly from the ground truth value $\theta^*$. This is typically achieved by using a two-sided test with the following null hypothesis:
\begin{equation}
H_{0}: \bar{\theta} = \theta^*, \quad H_{1}: \bar{\theta} \neq \theta^*.
\label{eq:trad_null}
\end{equation}
The corresponding t-statistic is given by
\begin{equation}
    t = \frac{\bar{\theta} - \theta^*}{s / \sqrt{N}}\, = \sqrt{N-1} \cdot \frac{\text{Bias}}{\sqrt{\text{Var}}}\,.
    \label{trad_t}
\end{equation}
where $\bar{\theta}$ is the sample mean across $N$ independent runs, and $s = \sqrt{\frac{1}{N-1} \sum_{i=1}^{N} (\hat{\theta}_i - \bar{\theta})^2}$ is the sample standard deviation. This shows that the traditional hypothesis test evaluates the relative magnitude (ratio) of the bias with respect to the variability of the estimator.

The corresponding $p$-value is \(p = 2 \cdot P(T_{N-1} > |t|)\), where $T_{N-1}$ follows a t-distribution with $N-1$ degrees of freedom. A small $p$-value ($p < \alpha$) indicates strong evidence against the null hypothesis, suggesting that the estimator’s mean is significantly different from the ground truth. Conversely, a large p-value ($p \geq \alpha$) implies insufficient evidence to reject the null hypothesis.

\paragraph{Limitations of Traditional Two-Sided Tests}
While the two-sided test provides a principled statistical framework, it suffers from critical limitations under variance reduction. Recall that variance naturally decreases with larger labeling budgets or through variance-reduction strategies, while bias typically fluctuates only mildly. Because the t-statistic reflects the ratio of bias to variance, a smaller variance shrinks the denominator in Eqn~\eqref{trad_t}, causing the t-statistic to grow even when bias remains constant. This also triggers a fundamental problem that when variance decreases, the test becomes increasingly sensitive to arbitrarily small biases, eventually rejecting the null hypothesis even for practically negligible deviations from $\theta^*$.

This sensitivity creates a serious issue: statistical significance does not imply practical significance. With sufficiently low variance, even a bias of 0.001 (0.1\% error) can yield a highly significant $p$-value ($p < 0.001$), leading to rejection of an estimator that is, for all practical purposes, perfectly adequate. The test conflates `detectably different from zero' with `meaningfully different from zero', providing no mechanism to distinguish between trivial and substantial biases.

Figure~\ref{fig:limit_p_rmse_example} illustrates this problem. At budget 20, the AT estimator exhibits negligible bias in plot (a), with its mean estimate closely aligned with the ground truth $\theta^* = 0.695$. Yet plot (c) shows that AT is rejected with $p = 0.044$ (under $\alpha = 0.05$ threshold). This rejection occurs not because AT's error is large, but because AT employs a variance-reduction strategy that achieves much lower variance than RS. The reduced variance inflates the t-statistic, making a practically negligible bias significant in contrast with a small variance. Meanwhile, RS passes the test despite its substantially higher variance, primarily because the inflated yet large denominator in the $t$-statistic overwhelms the contributions of the error.

\section{Fault-Tolerant Evaluation}
\label{sec:method}

\subsection{Overview of the Framework} 
To address the limitations of traditional metrics, we propose a fault-tolerant evaluation framework that explicitly accounts for both bias and variance within an application-specific tolerance level $\varepsilon$. Rather than testing whether the estimator is perfectly unbiased, our framework tests whether the estimator falls within a practically acceptable range around the ground truth.

We employ two one-sided hypothesis tests that evaluate whether our estimates are statistically consistent with the fault-tolerant approximation of the ground truth. We define two fault-tolerant null hypotheses ($H$) for the lower bound ($L$),
\begin{equation}
H_0^{(L)}: \bar{\theta} - \theta^* \leq -\varepsilon, \quad H_1^{(L)}: \bar{\theta} - \theta^* > -\varepsilon,
\end{equation}
and for the upper bound ($U$),
\begin{equation}
H_0^{(U)}: \bar{\theta} - \theta^* \geq \varepsilon, \quad H_1^{(U)}: \bar{\theta} - \theta^* < \varepsilon,
\end{equation}
where $\bar{\theta} = \frac{1}{N}\sum_{i=1}^{N} \hat{\theta}_i$ is the mean performance estimate across $N$ independent runs. The null hypotheses \(H_0^{(L)}, H_0^{(U)}\) state that $\bar{\theta}$ falls out of the fault-tolerant interval $[\theta^*-\epsilon, \theta^*+\epsilon]$, while the alternative hypotheses \(H_1^{(L)}, H_1^{(U)}\) state that $\bar{\theta}$ falls in it. 

\paragraph{Calculation of p-value} 
We conduct two \textit{one-sided} $t$-tests and obtain two $p$-values based on the distribution of estimates obtained by $N$ independent experiments at each $|\mathcal{S}|$. The t-statistics for both bounds are
\begin{equation}
t^{(L)} = \frac{\bar{\theta} - (\theta^* - \varepsilon)}{s / \sqrt{N}}, \quad
t^{(U)} = \frac{\varepsilon - (\bar{\theta} - \theta^*)}{s / \sqrt{N}}.
\end{equation}
The p-values for null hypotheses of both bounds are
\begin{equation}
    p^{(L)} = P\big(T_{N-1} > t^{(L)}\big), \quad p^{(U)} = P\big(T_{N-1} > t^{(U)}\big),
\end{equation}
where $T_{N-1}$ denotes the t-distribution with $N-1$ degrees of freedom.
Given a significance level $\alpha$, we reject both null hypotheses only if
\begin{equation}
(p^{(L)} < \alpha) \;\wedge\; (p^{(U)} < \alpha) \;\;\Leftrightarrow\;\; \max\{p^{(L)}, p^{(U)} \} < \alpha.
\end{equation}
Rejecting both null hypotheses indicates that the result by the estimator falls within the fault-tolerant interval $[\theta^* - \varepsilon, \theta^* + \varepsilon]$, since 
$-\varepsilon < \bar{\theta} - \theta^*$ (from the lower bound) and $\bar{\theta} - \theta^* < \varepsilon$ (from the upper bound), and we have $ | \bar{\theta} - \theta^* | < \varepsilon$.

\subsection{Tolerance Level Bounds Bias and Variance}

Our fault-tolerant evaluation framework addresses the limitations of traditional metrics by allowing practitioners to specify tolerance $\varepsilon$ that bounds both bias and variance according to the applications. This prevents two failure modes by traditional metrics discussed above: (1) rejecting practically adequate estimators when variance is low (traditional hypothesis test sensitivity), and (2) accepting consistently biased estimators that merely suppress variance (MSE or RMSE). The following derivation demonstrates how this formulation effectively overcomes these challenges.

Rejecting both null hypotheses $H_0^{(L)}$ and $H_0^{(U)}$ when both corresponding $p$-values are less than $\alpha$ is equivalent to both one-sided t-statistics exceeding the critical value $t_{\alpha,N-1}$,
\begin{align}
t^{(L)} &= \frac{\bar{\theta} - (\theta^* - \varepsilon)}{s / \sqrt{N}} > t_{\alpha,N-1}, \\
t^{(U)} &= \frac{\varepsilon - (\bar{\theta} - \theta^*)}{s / \sqrt{N}} > t_{\alpha,N-1}.
\end{align}
Rearranging these inequalities yields
\begin{align}
\bar{\theta} - \theta^* + \varepsilon & > t_{\alpha,N-1} \cdot \frac{s}{\sqrt{N}}, \\
\varepsilon - (\bar{\theta} - \theta^*) & > t_{\alpha,N-1} \cdot \frac{s}{\sqrt{N}}.
\end{align}
Combining both conditions yields the following requirement for rejecting both null hypotheses (\ie concluding the estimator falls within the FT interval)
\begin{equation}
|\bar{\theta} - \theta^*| + t_{\alpha,N-1} \cdot \frac{s}{\sqrt{N}} < \varepsilon.
\label{eq:eps_bound}
\end{equation}
Substituting the bias and variance expressions, we obtain
\begin{equation}
|\text{Bias}| + t_{\alpha,N-1} \cdot \sqrt{\frac{\text{Var}}{N-1}} < \varepsilon.
\label{eps_relation}
\end{equation}
This decomposition shows that $\varepsilon$ explicitly bounds the combined effect of bias and variance. By calibrating $\varepsilon$ appropriately, it can effectively mitigate the low-variance pitfalls in both RMSE and $p$-value based evaluations.

\subsection{Practical Guidance}
\label{sec:prac_guide}
Selecting an appropriate tolerance level $\varepsilon$ is critical for balancing statistical rigor with practical relevance. To avoid requiring users to manually specify this tolerance, we propose Algorithm~\ref{algo:set_epsilon} that determines the hyperparameter for evaluation experiments. 

The algorithm employs a binary search to efficiently identify the minimum discrimination margin $\delta^*$ that distinguishes between the two estimators under \ourframework. As shown in Equation~\ref{eps_relation}, this margin controls the tolerated bias: a smaller $\delta$ imposes stricter requirements on estimator accuracy. For each candidate $\delta$, the algorithm iterates through labeling budgets from 1 to $|\mathcal{S}|$. For each budget $k$, the function $\text{Run}(E, N, k)$ executes estimator $E$ for $N$ independent trials, returning the sample mean $\bar{\theta}_k$ and standard deviation $s_k$ of the resulting estimates. The tolerance level $\varepsilon_{k}$ is then dynamically computed for each estimator based on its $s_k$. This adaptive tolerance ensures that the evaluation remains appropriately calibrated across all budget levels: at low budgets, larger tolerances accommodate higher sampling uncertainty, while at high budgets, tighter tolerances ensure that persistent bias is detected.

If the two estimators yield different FT-Eval outcomes at any budget $k$, the algorithm records the current $\delta$ and continues searching for a smaller margin. Finally, the algorithm returns the smallest discrimination margin $\delta^*$ at which the estimators can be distinguished, or \texttt{Null} if the algorithm cannot find a margin in $[0, 1]$ separates their behavior.



\section{Experimental Settings}

\subsection{Dataset and Model}
For black-box classifiers, we follow the experimental setup from \citet{ji2021active}, which introduces an active Bayesian assessment framework for evaluating models \(\mathcal{M}\) such as ResNet and BERT across different domains. Our experiments cover three computer vision datasets (\ie CIFAR-100 \cite{krizhevsky2009learning}, ImageNet \cite{deng2009imagenet}, and SVHN \cite{netzer2011reading}) as well as two natural language processing datasets (\ie 20 Newsgroups \cite{lang1995newsweeder} and DBpedia \cite{auer2007dbpedia}). 

For LLMs, we use the MMLU-Pro benchmark \cite{wang2024mmlu}, which contains 14 disciplinary subsets and 12,032 multiple-choice questions, each with ten answer options. Following the benchmark protocol, evaluation is conducted under a 5-shot Chain-of-Thought (CoT) prompting setup \cite{wei2022chain}. We focus on the five largest subsets (Math, Physics, Chemistry, Law, and Engineering) based on question count, with GPT-4o Mini \cite{menick2024gpt} serving as the target model \(\mathcal{M}\). Detailed dataset and model specifications are provided in the Appendix~\ref{sec:dataset_model}.

\begin{algorithm}[h!]
\caption{Auto-selecting discrimination margin $\delta$ and dynamic tolerance $\varepsilon$}
\label{algo:set_epsilon}
\begin{algorithmic}
\STATE {\bfseries Input:} Estimators $E_A, E_B$, ground truth $\theta^*$, target budget $|\mathcal{S}|$, number of runs $N$, significance level $\alpha$
\STATE {\bfseries Output:} Discrimination margin $\delta^*$ or \texttt{Null} if estimators are indistinguishable

\STATE $\delta^* \leftarrow \texttt{Null}$, $\text{low} \leftarrow 0$, $\text{high} \leftarrow 1$

\WHILE{$\text{high} - \text{low} \geq 0.01$}
    \STATE $\delta \leftarrow (\text{low} + \text{high}) / 2$, $\text{distinguished} \leftarrow \texttt{False}$
    \FOR{$k = 1$ {\bfseries to} $|\mathcal{S}|$}
        \STATE $\bar{\theta}_{A,k}, s_{A,k} \leftarrow \text{Run}(E_A, N, k)$
        \STATE $\bar{\theta}_{B,k}, s_{B,k} \leftarrow \text{Run}(E_B, N, k)$
        \STATE $\varepsilon_{A,k} \leftarrow \delta + t_{\alpha, N-1} \cdot \dfrac{s_{A,k}}{\sqrt{N}}$
        \STATE $\varepsilon_{B,k} \leftarrow \delta + t_{\alpha, N-1} \cdot \dfrac{s_{B,k}}{\sqrt{N}}$
        \STATE $\text{pass}_A \leftarrow \text{FT-Eval}(\bar{\theta}_{A,k}, s_{A,k}, \theta^*, \varepsilon_{A,k}, N, \alpha)$
        \STATE $\text{pass}_B \leftarrow \text{FT-Eval}(\bar{\theta}_{B,k}, s_{B,k}, \theta^*, \varepsilon_{B,k}, N, \alpha)$
        
        \IF{$\text{pass}_A \neq \text{pass}_B$}
            \STATE $\text{distinguished} \leftarrow \texttt{True}$
            \STATE \textbf{break}
        \ENDIF 
    \ENDFOR
    
    \IF{$\text{distinguished}$} 
        \STATE $\delta^* \leftarrow \delta$, $\text{high} \leftarrow \delta$ \hfill$\triangleright$ Search left for smaller $\delta$
    \ELSIF{$\text{pass}_A \land \text{pass}_B$}
        \STATE $\text{high} \leftarrow \delta$ \hfill$\triangleright$ Both pass; search left
    \ELSE
        \STATE $\text{low} \leftarrow \delta$ \hfill$\triangleright$ Both fail; search right
    \ENDIF
\ENDWHILE
\STATE \textbf{return} $\delta^*$ 
\end{algorithmic}
\end{algorithm}

\subsection{Active Testing and Random Sampling Estimators}
\label{sec:at_rs_est_setup}
We employ \textbf{two} distinct active testing configurations tailored to different model types. For black-box classifiers, we adopt the active testing estimator proposed by \citet{ji2021active}, maintaining their original settings including informed priors that leverage classifier confidence scores and predicted labels for dataset grouping. For LLMs, we develop an adapted version of this algorithm to address the unique challenges of evaluating LLMs, including the use of uninformed priors, task embedding-based partitioning, and density sampling strategies (details see Appendix \ref{AT_LLM}).

In both experimental settings, we evaluate the active testing approach against a random sampling estimator, which selects tasks uniformly at random from the available pool for annotation and evaluation. This method follows prior studies in active testing \cite{kossen2021active, ji2021active, huang2024active}. Specifically, for each target model $\mathcal{M}$, we vary the labeling budget size $|\mathcal{S}| \in \{1, \dots, 100\}$ and sample subsets $\mathcal{S} \subseteq \mathcal{D}$ to be labeled and evaluated. At each budget level, we conduct $N = 100$ independent runs with different random seeds and assess estimators using both traditional metrics (\eg RMSE, $p$-value) and \ourframework.

\section{Experimental Results}

\subsection{Conflicts in Traditional Evaluation}
\label{sec:conflict_signal}

\begin{figure*}[h!]
  \centering
  \includegraphics[width=\linewidth]{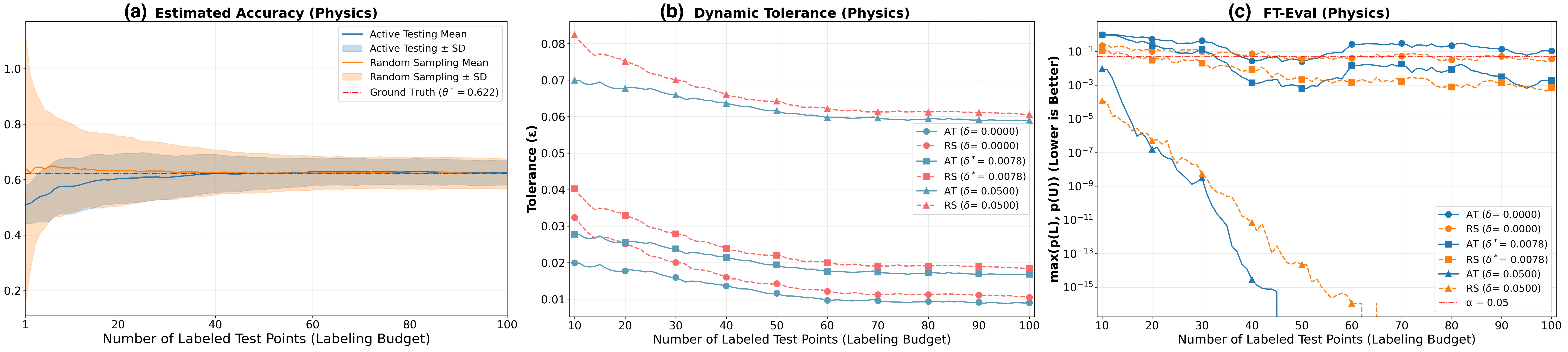}
  \caption{Comparison of Active Testing and Random Sampling estimators on MMLU-Pro (Physics): (a) Estimated Accuracy across labeling budgets, (b) Dynamic Tolerance $\varepsilon$ under different Discrimination Margins $\delta$, (c) FT-Eval, measured as $\max(p^{(L)}, p^{(U)})$.}
  \label{fig:ft-eval}
  \vspace*{-0.5em}
\end{figure*}

Table~\ref{tab:p-value:rmse} reveals a significant inconsistency: $p$-values and RMSE often provide conflicting guidance on which estimator performs better, calling into question their reliability as standalone evaluation criteria. Across 30 experimental settings (10 datasets $\times$ 3 labeling budgets), the two metrics disagree in 73.3\% (22 out of 30) cases. Notably, for datasets such as ImageNet, SVHN, 20 Newsgroup, DBpedia, and Physics, $p$-values and RMSE diverge under \textbf{all} three labeling budgets shown here. A consistent pattern emerges: AT frequently produces extremely small $p$-values, suggesting strong statistical evidence against the null hypothesis (Equation~\ref{eq:trad_null}), yet simultaneously achieves lower RMSE than RS. In other words, AT appears statistically `invalid' under the $p$-value criterion while being more accurate under RMSE.

\begin{table}[h!]
\centering
\caption{Comparison of Active Testing (AT) and Random Sampling (RS) estimators across tasks and labeling budgets. Reported metrics include p-value ($\uparrow$ higher is better) and RMSE ($\downarrow$ lower is better). Color legend: {\color{green!20}\rule{7pt}{7pt}} better value, {\color{red!20}\rule{7pt}{7pt}} worse value.}
\resizebox{\columnwidth}{!}{%
\footnotesize
\begin{tabular}{l@{\hskip -0.2em}ccccc}
\toprule
\multirow{2}{*}{\textbf{Dataset \(\mathcal{D}\)}} & 
\multirow{2}{*}{\makecell{\textbf{Labelling} \\ \textbf{Budget \(\mathcal{|S|}\)}}} & 
\multicolumn{2}{c}{\textbf{p-value ($\uparrow$)}} & 
\multicolumn{2}{c}{\textbf{RMSE ($\downarrow$)}} \\
\cmidrule(l){3-4} \cmidrule(l){5-6}
& & AT & RS & AT & RS \\
\midrule
\multirow{3}{*}{CIFAR-100} 
& 20  & \cellcolor{red!20}{2.7e-112} & \cellcolor{green!20}{1.000} & \cellcolor{green!20}{0.089} & \cellcolor{red!20}{0.098} \\
& 60  & \cellcolor{red!20}{4.3e-82} & \cellcolor{green!20}{0.490} & \cellcolor{red!20}{0.075} & \cellcolor{green!20}{0.055} \\
& 80 & \cellcolor{red!20}{3.3e-76} & \cellcolor{green!20}{0.620} & \cellcolor{red!20}{0.074} & \cellcolor{green!20}{0.050} \\
\cmidrule{1-6}
\multirow{3}{*}{ImageNet} 
& 20  & \cellcolor{red!20}{2.0e-178} & \cellcolor{green!20}{0.177} & \cellcolor{green!20}{0.043} & \cellcolor{red!20}{0.085} \\
& 60  & \cellcolor{red!20}{1.1e-160} & \cellcolor{green!20}{0.973} & \cellcolor{green!20}{0.042} & \cellcolor{red!20}{0.049} \\
& 80 & \cellcolor{red!20}{1.8e-146} & \cellcolor{green!20}{0.830} & \cellcolor{green!20}{0.042} & \cellcolor{red!20}{0.046} \\
\cmidrule{1-6}
\multirow{3}{*}{SVHN} 
& 20  & \cellcolor{red!20}{3.6e-04} & \cellcolor{green!20}{0.691} & \cellcolor{green!20}{0.025} & \cellcolor{red!20}{0.062} \\
& 60  & \cellcolor{red!20}{5.2e-09} & \cellcolor{green!20}{0.376} & \cellcolor{green!20}{0.029} & \cellcolor{red!20}{0.039} \\
& 80 & \cellcolor{red!20}{1.1e-09} & \cellcolor{green!20}{0.881} & \cellcolor{green!20}{0.028} & \cellcolor{red!20}{0.033} \\
\cmidrule{1-6}
\multirow{3}{*}{20 Newsgroup} 
& 20  & \cellcolor{red!20}{0.044} & \cellcolor{green!20}{0.495} & \cellcolor{green!20}{0.034} & \cellcolor{red!20}{0.109} \\
& 60  & \cellcolor{red!20}{1.9e-07} & \cellcolor{green!20}{0.839} & \cellcolor{green!20}{0.036} & \cellcolor{red!20}{0.057} \\
& 80 & \cellcolor{red!20}{1.0e-08} & \cellcolor{green!20}{0.745} & \cellcolor{green!20}{0.034} & \cellcolor{red!20}{0.050} \\
\cmidrule{1-6}
\multirow{3}{*}{DBpedia} 
& 20  & \cellcolor{red!20}{4.6e-28} & \cellcolor{green!20}{0.823} & \cellcolor{green!20}{0.008} & \cellcolor{red!20}{0.022} \\
& 60  & \cellcolor{red!20}{5.1e-38} & \cellcolor{green!20}{0.999} & \cellcolor{green!20}{0.007} & \cellcolor{red!20}{0.014} \\
& 80 & \cellcolor{red!20}{1.1e-41} & \cellcolor{green!20}{0.673} & \cellcolor{green!20}{0.007} & \cellcolor{red!20}{0.012} \\
\cmidrule{1-6}
\multirow{3}{*}{Math} 
& 20  & \cellcolor{red!20}{4.6e-08} & \cellcolor{green!20}{0.684} & \cellcolor{red!20}{0.096} & \cellcolor{green!20}{0.090} \\
& 60  & \cellcolor{red!20}{2.2e-04} & \cellcolor{green!20}{0.783} & \cellcolor{red!20}{0.059} & \cellcolor{green!20}{0.047} \\
& 80 & \cellcolor{red!20}{0.001} & \cellcolor{green!20}{0.694} & \cellcolor{red!20}{0.050} & \cellcolor{green!20}{0.045} \\
\cmidrule{1-6}
\multirow{3}{*}{Physics} 
& 20  & \cellcolor{red!20}{0.040} & \cellcolor{green!20}{0.479} & \cellcolor{green!20}{0.091} & \cellcolor{red!20}{0.126} \\
& 60  & \cellcolor{red!20}{0.173} & \cellcolor{green!20}{0.828} & \cellcolor{green!20}{0.049} & \cellcolor{red!20}{0.061} \\
& 80 & \cellcolor{red!20}{0.222} & \cellcolor{green!20}{0.912} & \cellcolor{green!20}{0.047} & \cellcolor{red!20}{0.056} \\
\cmidrule{1-6}
\multirow{3}{*}{Chemistry} 
& 20  & \cellcolor{green!20}{0.865} & \cellcolor{red!20}{0.322} & \cellcolor{green!20}{0.098} & \cellcolor{red!20}{0.115} \\
& 60  & \cellcolor{red!20}{0.050} & \cellcolor{green!20}{0.084} & \cellcolor{green!20}{0.053} & \cellcolor{red!20}{0.061} \\
& 80 & \cellcolor{green!20}{0.157} & \cellcolor{red!20}{0.073} & \cellcolor{green!20}{0.048} & \cellcolor{red!20}{0.058} \\
\cmidrule{1-6}
\multirow{3}{*}{Law} 
& 20  & \cellcolor{red!20}{0.271} & \cellcolor{green!20}{0.993} & \cellcolor{green!20}{0.097} & \cellcolor{red!20}{0.114} \\
& 60  & \cellcolor{red!20}{0.156} & \cellcolor{green!20}{0.320} & \cellcolor{green!20}{0.062} & \cellcolor{red!20}{0.071} \\
& 80 & \cellcolor{green!20}{0.378} & \cellcolor{red!20}{0.304} & \cellcolor{green!20}{0.055} & \cellcolor{red!20}{0.058} \\
\cmidrule{1-6}
\multirow{3}{*}{Engineering} 
& 20  & \cellcolor{red!20}{0.384} & \cellcolor{green!20}{0.570} & \cellcolor{green!20}{0.093} & \cellcolor{red!20}{0.096} \\
& 60  & \cellcolor{green!20}{0.950} & \cellcolor{red!20}{0.185} & \cellcolor{red!20}{0.056} & \cellcolor{green!20}{0.055} \\
& 80 & \cellcolor{green!20}{0.646} & \cellcolor{red!20}{0.472} & \cellcolor{red!20}{0.048} & \cellcolor{green!20}{0.045} \\
\bottomrule
\end{tabular}
}
\label{tab:p-value:rmse}
\vspace{-0.5em}
\end{table}

This contradiction stems from the inherent limitations of both metrics, as discussed in Sections~\ref{sec:limit_trad}. The $t$-statistic underlying the $p$-value reflects the ratio of estimator bias to variance. With lower variance, $p$-values become more sensitive, rejecting the null even for negligible deviations from $\theta^*$. RS, being an unbiased but high-variance estimator, frequently achieves high $p$-values because its large variance inflates the denominator of the test statistic, masking its error magnitude. AT, by contrast, reduces variance through selective sampling, which amplifies the relative effect of even small biases, driving small $p$-values. Meanwhile, RMSE tells a different story. Because RMSE decomposes into bias and variance, its value is dominated by whichever component is larger. For RS at low budgets, the large variance component dominates, resulting in high RMSE despite low bias. For AT, the substantially lower variance compensates for its slightly higher bias, leading to a smaller RMSE overall. Taken together, these conflicting signals demonstrate that $p$-values and RMSE systematically misrepresent estimator quality under different bias–variance regimes. 

\subsection{Analysis of FT-Eval}

To demonstrate how \ourframework addresses the disadvantages of traditional metrics, and how Algorithm~\ref{algo:set_epsilon} manage to select appropriate discrimination margin $\delta$ and tolerance level $\varepsilon$, we examine the MMLU-Pro (Physics) dataset as a representative case study comparing AT and RS estimators.

Figure~\ref{fig:ft-eval} (a) presents the accuracy estimates on the Physics dataset. RS exhibits high variance at low labeling budgets but gradually converges to the ground-truth accuracy as the budget increases. In contrast, AT starts near 0.5 due to initialization with uninformed priors and improves steadily with lower variance than RS. This behavior leads to conflicting RMSE and $p$-value assessments, as reported in Table~\ref{tab:p-value:rmse}.



FT-Eval resolves this inconsistency by incorporating a dynamic tolerance $\varepsilon$. Figure~\ref{fig:ft-eval}(b) shows the evolution of $\varepsilon$ under three discrimination margins $\delta$, where $\delta^*$ denotes the optimal value selected by Algorithm~\ref{algo:set_epsilon}. Smaller values of $\delta$ induce smaller tolerance, corresponding to stricter FT tests. Furthermore, there is a consistent trend in which the $\varepsilon$ assigned to the AT estimator is smaller than that of RS at early budgets, and both tolerances gradually decrease and converge as the labeling budget increases. This behavior reflects the lower variance of AT in the early stage and the variance decay of both estimators with increasing labeled points, as expected under the law of large numbers~\cite{kolmogorov1963theory}. By integrating the dynamic tolerance $\varepsilon$, FT-Eval ensures a fair and comparable evaluation for both estimators across all budget levels.

Figure~\ref{fig:ft-eval}(c) reports the FT-Eval results and highlights the importance of automatically selecting $\delta^*$. When $\delta = 0.0$, the resulting tolerance $\varepsilon$ is overly restrictive, causing both estimators to fail the FT test across nearly all labeling budgets and preventing meaningful comparison. Conversely, when $\delta = 0.05$, the tolerance becomes too permissive, allowing both estimators to pass the FT test from the beginning. The automatically selected margin $\delta^* = 0.0078$ strikes an effective balance. It enables clear discrimination between estimators while remaining as small as possible. Under this setting, RS passes the FT test earlier than AT. Furthermore, by accounting for variance decay through dynamic tolerance, FT test correctly rejects AT by a budget of 35 samples due to its systematic bias, which RMSE fails to capture.



\section{Conclusions}
In summary, we address a critical challenge in reliably evaluating sample-efficient performance estimators. Our analysis reveals fundamental flaws in traditional metrics under low variance conditions, where they yield conflicting conclusions in 73\% of cases. The proposed \ourframework resolves these issues by explicitly bounding both bias and variance within adjustable tolerance levels, enabling consistent and interpretable assessments across different variance regimes. We further introduce an algorithm that automatically selects the tolerance by identifying the smallest discrimination margin that distinguishes between competing estimators.



\newpage

\section*{Impact Statement}



This work aims to advance the field of machine learning by improving the reliability of performance evaluation for sample-efficient estimators. By providing a fault-tolerant evaluation framework, our approach has the potential to support more robust model assessment in settings where labeling resources are limited, such as API-based AI services and large-scale agentic AI systems. More reliable evaluation can lead to better-informed model selection, reduced deployment risk, and more efficient use of human annotation effort.

The proposed framework is methodological in nature and does not introduce new model architectures or decision-making policies. As such, it is not expected to directly cause negative societal impacts. However, as with any evaluation methodology, its use in downstream applications should be guided by domain-specific considerations, including fairness, accountability, and responsible deployment practices. We encourage practitioners to apply the framework in conjunction with appropriate safeguards when used in high-stakes or sensitive settings.

\section*{Acknowledgements}
Zihan Zhu was supported by the Digital Finance Cooperative Research Centre (DFCRC) through a DFCRC industry scholarship. The whole team was supported by 2024 FSE Strategic Startup.


\bibliography{references}
\bibliographystyle{icml2026}

\newpage
\appendix
\onecolumn
\section{Dataset \& Model}
\label{sec:dataset_model}
Table~\ref{tab:blackbox_data} and \ref{tab:mmlu_data} summarize the representative datasets, the corresponding models, and their ground-truth accuracies \(\theta^*\) used in our evaluation.

\begin{table}[h!]
\centering
\caption{Datasets, prediction models, and ground-truth accuracies used in the black-box classifier experiments.}
\vspace{-1mm}
\begin{tabular}{lccc}
\toprule
\textbf{Dataset \(\mathcal{D}\)} & \textbf{Size \(|\mathcal{D}|\)} & \textbf{Model \(\mathcal{M}\)} & \textbf{Accuracy $\theta^*$} \\
\midrule
CIFAR-100  & 10K   & ResNet-110   & 0.728 \\
ImageNet & 50K   & ResNet-152   & 0.787 \\
SVHN       & 26K   & ResNet-152   & 0.904 \\
20 Newsgroup    & 7.5K  & BERT\textsubscript{BASE}  & 0.695 \\
DBpedia    & 70K   & BERT\textsubscript{BASE}  & 0.990 \\
\bottomrule
\end{tabular}
\label{tab:blackbox_data}
\vspace{-3.5mm}
\end{table}

\begin{table}[h!]
    \centering
    \caption{Top five MMLU-Pro disciplines and GPT-4o mini ground-truth accuracies used in the LLM experiments.}
    \vspace{-1mm}
    \begin{tabular}{lcc}
        \toprule
        \textbf{Dataset \(\mathcal{D}\)} & \textbf{Size \(|\mathcal{D}|\)} & \textbf{Accuracy \(\theta^*\)} \\
        \midrule
        Math         & 1351 & 0.737 \\
        Physics      & 1299 & 0.622 \\
        Chemistry    & 1132 & 0.633 \\
        Law         & 1101 & 0.372 \\
        Engineering  & 969  & 0.387 \\
        \bottomrule
    \end{tabular}
    \label{tab:mmlu_data}
    \vspace{-3.5mm}
\end{table}

\section{Active Bayesian Assessment of LLM}
\label{AT_LLM}

Our LLM-adapted active testing algorithm (shown in Algorithm \ref{algo1}) addresses two major challenges that distinguish LLM evaluation from traditional approaches: (1) \textbf{Evaluation Costs}: running LLMs on thousands of prompts is computationally intensive; and (2) \textbf{Labeling Costs}: constructing such large-scale benchmarks requires substantial time and manual efforts for labeling.

\begin{algorithm}[h!]
\caption{Active Bayesian Assessment of LLM}
\label{algo1}
\begin{algorithmic}
\STATE {\bfseries Input:} Dataset $\mathcal{D} = \{D_1, D_2, D_3, \dots, D_n\}$, LLM model $\mathcal{M}$, labeling budget $|\mathcal{S}|$, groups $g \in [1, G]$
\STATE {\bfseries Output:} Maximum a posteriori (MAP) global estimate $\hat{\theta}_\mathcal{D}$
\STATE Initialize priors: $\alpha_{g} = 0.5$, $\beta_{g} = 0.5$ for all $g$
\FOR{$i = 1$ {\bfseries to} $|\mathcal{S}|$}
    \STATE // Sample parameters for each group
    \STATE $\hat{\theta}_{g} \sim \text{Beta}(\alpha_{g}, \beta_{g})$
    \STATE // Select a group by maximizing expected reward
    \STATE $\widetilde{g} = \arg\max_{g} p_g \cdot \big[\text{Var}(\hat{\theta}_{g} \mid \mathcal{L}) - \big(\hat{\theta}_{g} \cdot \text{Var}(\hat{\theta}_{g} \mid \mathcal{L}, z=1) + (1 - \hat{\theta}_{g}) \cdot \text{Var}(\hat{\theta}_{g} \mid \mathcal{L}, z=0) \big) \big]$
    \STATE // Select a task from group $\widetilde{g}$ using density sampling
    \STATE $D_i \leftarrow \text{DensitySample}(\widetilde{g})$
    \STATE // Query true label and verify model $\mathcal{M}$ prediction
    \STATE $z_i \leftarrow \mathbb{I}(Y_i, \hat{Y}_i)$, where $z_i \in \{0,1\}$
    \STATE // Update parameters of group $\widetilde{g}$
    \STATE $\alpha_{\widetilde{g}} \leftarrow \alpha_{\widetilde{g}} + z_i$
    \STATE $\beta_{\widetilde{g}} \leftarrow \beta_{\widetilde{g}} + 1 - z_i$
\ENDFOR
\STATE // Compute global estimate
\STATE $\hat{\theta}_{\mathcal{D}} = \sum_{g=1}^{G} p_g \cdot \frac{\alpha_g}{\alpha_g + \beta_g}$
\STATE {\bfseries return} $\hat{\theta}_\mathcal{D}$
\end{algorithmic}
\end{algorithm}

To tackle these issues, we make three key modifications to the original algorithm. First, we adjust the grouping method by applying semantic clustering based on OpenAI’s embedding model (\texttt{text-embedding-3-small}), rather than confidence-based partitioning. The optimal number of groups is determined via silhouette score to capture meaningful task similarities. Second, we use uninformative Beta(0.5, 0.5) priors for all groups, avoiding the need to infer over the entire test set and allowing the algorithm to learn purely from labeled observations. Third, we implement density sampling within each selected group to prioritize representative tasks while maintaining diversity, thereby avoiding both outliers and uniform sampling inefficiencies. These modifications significantly reduce computational overhead by eliminating the need for full dataset inference during active selection, making the approach practical for large-scale LLM evaluation where inference costs are substantial.

\section{Additional Experimental Results}
\label{app:add_exp}

\begin{figure*}[h!]
    \centering
    \includegraphics[width=\linewidth]{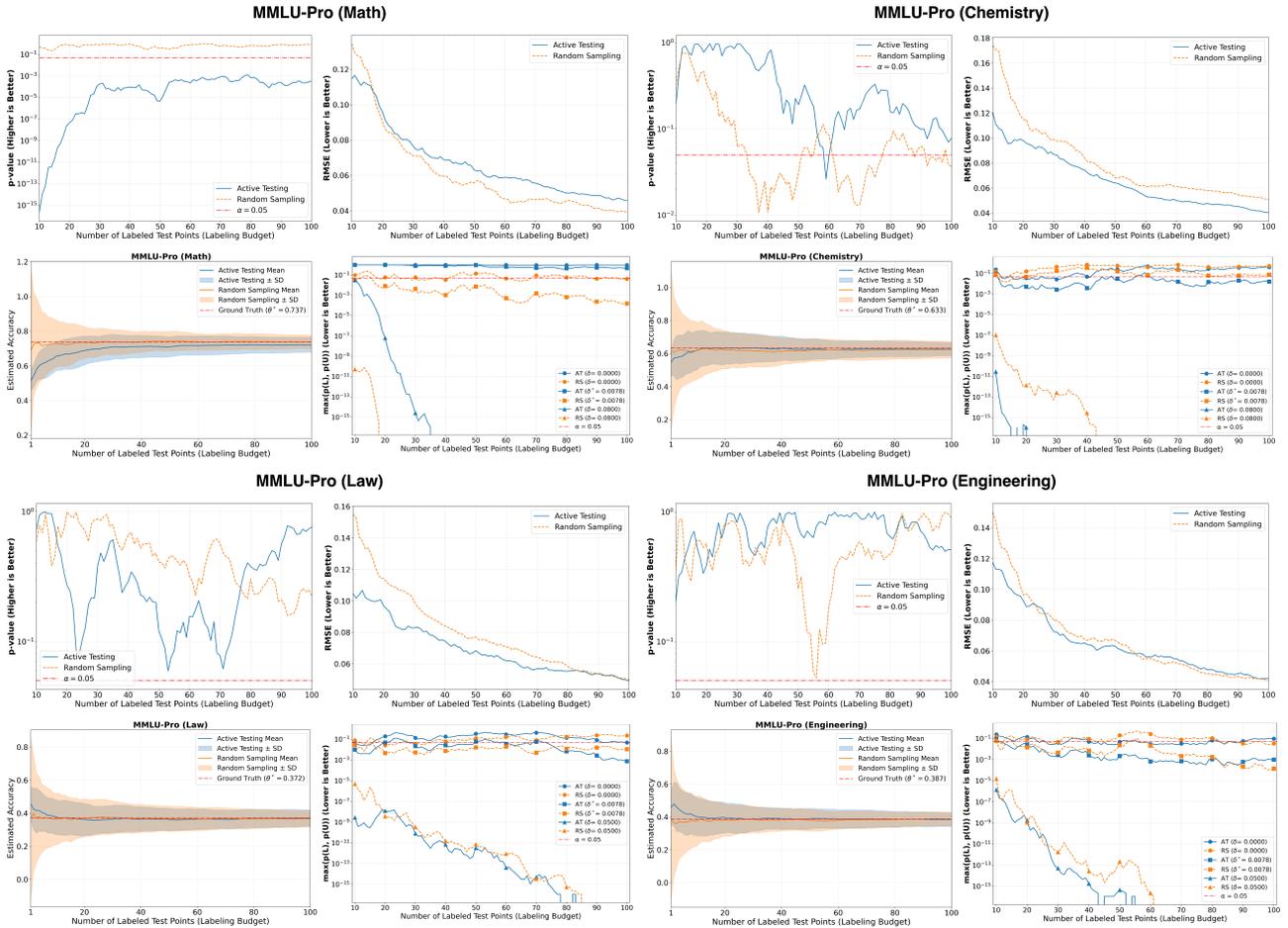}
    \caption{Comparison of accuracy estimates for Active Testing (AT) and Random Sampling (RS) estimators on MMLU-Pro (Math, Chemistry, Law, Engineering) Datasets using three metrics: $p$-value, RMSE, and FT-Eval (measured as $\max(p^{(L)}, p^{(U)})$).}
    \label{fig:app_exp_llm}
\end{figure*}
\begin{figure*}[h!]
    \centering
    \includegraphics[width=\linewidth]{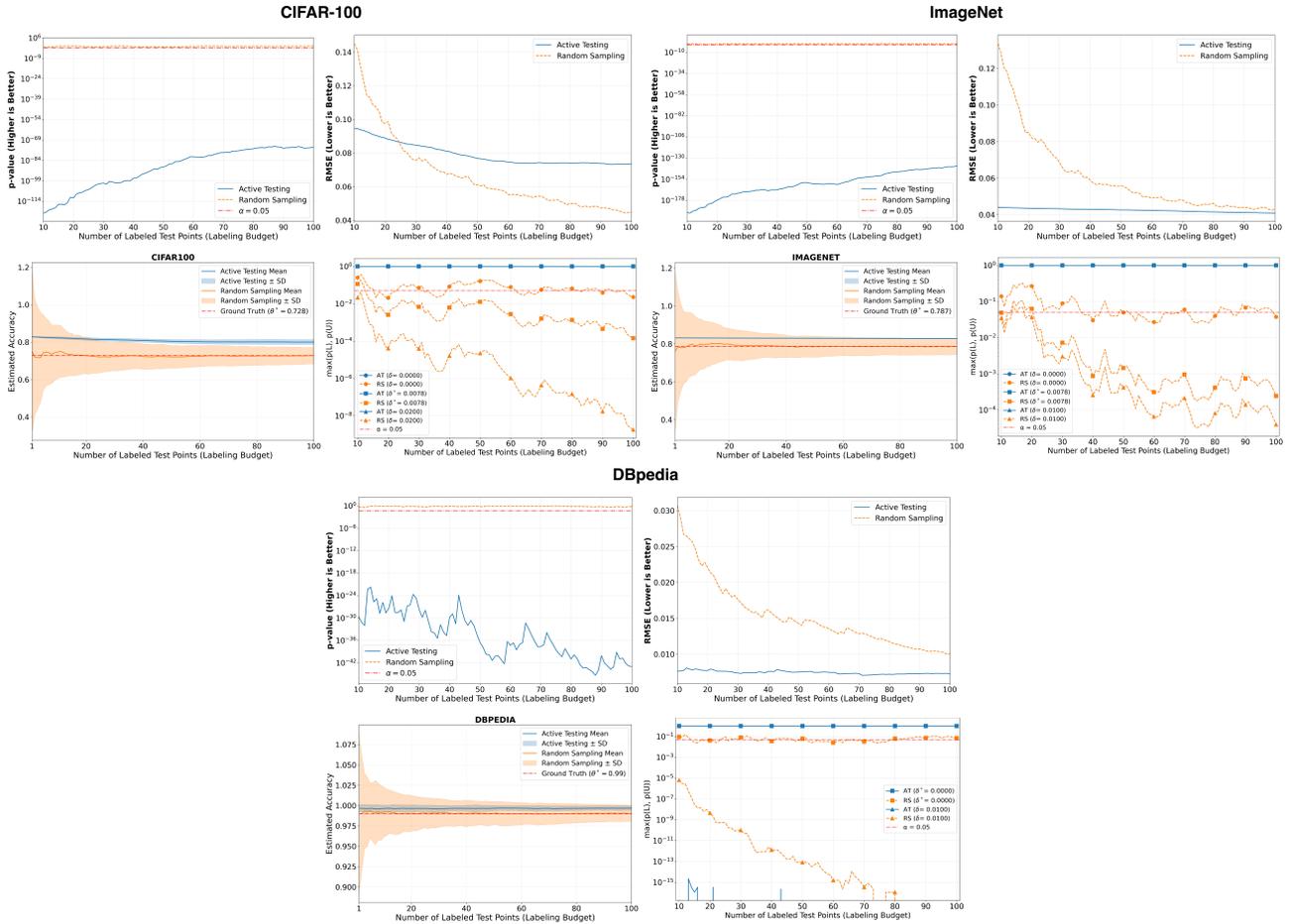}
    \caption{Comparison of accuracy estimates for Active Testing (AT) and Random Sampling (RS) estimators on CIFAR-100, ImageNet, DBpedia Datasets using three metrics: $p$-value, RMSE, and FT-Eval (measured as $\max(p^{(L)}, p^{(U)})$).}
    \label{fig:app_exp_classifier}
\end{figure*}

To provide comprehensive validation of our fault-tolerant evaluation (\ourframework), we present complete experimental results across all remaining datasets in our evaluation suite. Figures~\ref{fig:app_exp_llm} and~\ref{fig:app_exp_classifier} compare traditional metrics ($p$-value, RMSE) with the proposed \ourframework metric ($\max(p^{(L)}, p^{(U)})$) under dynamic tolerance levels with different discrimination margins $\delta$. Across all datasets, \ourframework with an appropriately selected $\delta$ yields more coherent and interpretable assessments than traditional metrics, supporting the findings from our main experiments.

\end{document}

%% file: def.tex
\usepackage{stmaryrd}
\usepackage{bbm}
\usepackage{xspace}
\usepackage[table]{xcolor}
\usepackage{todonotes}
\usepackage{multirow}
\usepackage{graphicx}
\usepackage[most]{tcolorbox}
\usepackage{makecell}

\def\ourframework{\textbf{FT-Eval}\xspace}

\def\eg{{\em e.g.,}\xspace}
\def\ie{{\em i.e.,}\xspace}